\pdfoutput=1
\documentclass[11pt]{article}

\usepackage{acl}

\usepackage{times}
\usepackage{latexsym}
\usepackage{enumitem}
\usepackage{graphicx}
\usepackage{amsmath}
\usepackage{hyperref}
\usepackage{algorithm}
\usepackage{algorithmic}
\usepackage{amssymb}
\usepackage{multirow}
\usepackage{booktabs}

\usepackage[T1]{fontenc}
\usepackage[utf8]{inputenc}

\usepackage{microtype}
\title{Pre-training to Match for Unified Low-shot Relation Extraction}

\author{Fangchao Liu${}^{1,3}$,
    Hongyu Lin${}^{1,}$\thanks{~ Corresponding authors.}\ \  , Xianpei Han${}^{1,2,4}$, Boxi Cao${}^{1,3}$, Le Sun${}^{1,2}$ \\
  ${}^{1}$Chinese Information Processing Laboratory ~ ${}^{2}$State Key Laboratory of Computer Science \\
  Institute of Software, Chinese Academy of Sciences, Beijing, China\\
  ${}^{3}$University of Chinese Academy of Sciences, Beijing, China \\
  ${}^{4}$Beijing Academy of Artificial Intelligence, Beijing, China \\
   {\tt \{fangchao2017,hongyu,xianpei,boxi2020,sunle\}@iscas.ac.cn}
 }

\date{}

\begin{document}
\maketitle
\begin{abstract}

    Low-shot relation extraction~(RE) aims to recognize novel relations with very few or even no samples, which is critical in real scenario application. Few-shot and zero-shot RE are two representative low-shot RE tasks, which seem to be with similar target but require totally different underlying abilities. In this paper, we propose Multi-Choice Matching Networks to unify low-shot relation extraction. To fill in the gap between zero-shot and few-shot RE, we propose the triplet-paraphrase meta-training, which leverages triplet paraphrase to pre-train zero-shot label matching ability and uses meta-learning paradigm to learn few-shot instance summarizing ability. Experimental results on three different low-shot RE tasks show that the proposed method outperforms strong baselines by a large margin, and achieve the best performance on few-shot RE leaderboard\footnote{\href{https://thunlp.github.io/2/fewrel2\_nota.html}{https://thunlp.github.io/2/fewrel2\_nota.html}}. 

\end{abstract}

\section{Introduction}
Relation extraction (RE) aims to extract the relation between two given entities in the context. The most popular approaches to build RE models are based on supervised learning~\citep{zeng-2014, soares-2019}. Despite the superior performance, supervised relation extraction approaches severely suffer from the data bottleneck, which restricts their application to more relation types in real scenarios.

Consequently, low-shot relation extraction has become a recent research hotspot in RE area. There are two mainstream learning paradigms widely explored in low-shot relation extraction, namely zero-shot RE~\citep{levy-2017-zero} and few-shot RE~\citep{han-2018-fewrel}. 
Few-shot relation extraction aims to identify instances of novel relation type with only a few illustrative instances, while zero-shot RE is more progressive, which only uses external knowledge and the name or definition of the novel relations to recognize them. Because low-shot RE only requires very limited manually annotated data, it can effectively alleviate data bottlenecks in conventional RE and therefore attached great attention.
\begin{figure}[!t]
    \setlength{\belowcaptionskip}{-0.6cm}
    \centering
    \includegraphics[width=0.4\textwidth]{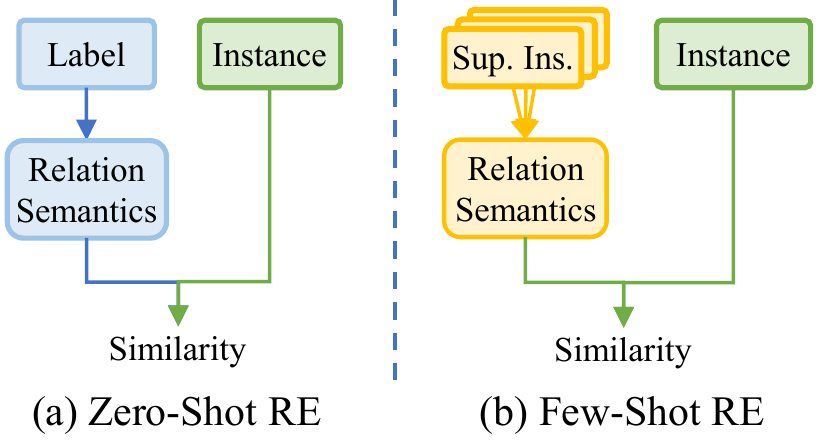}
    \caption{Difference between zero-shot RE and few-shot RE. (a) Zero-shot requires for the ability of label semantic matching, while (b) Few-shot requires for the ability of support instance (Sup. Ins.) summarizing.}
    \label{fig:compare}
\end{figure}

However, even with similar goals, zero-shot RE and few-shot RE actually require different fundamental abilities.
Specifically, zero-shot RE is built on \emph{label semantic matching} ability, which requires models to sufficiently exploit the label semantic of given novel relations, and matches relations and queries based on their underlying semantics. While few-shot RE is built on \emph{instance semantic summarizing} ability, which requires a model to quickly generalize to novel relations by summarizing critical information from few-shot instances. Due to this fundamental difference, current state-of-the-art architectures are separately learned to deal with these two low-shot RE tasks. For zero-shot RE, the most popular solution is to transform it into a textual entailment~\cite{obamuyide-2018-zero,sainz-2021-entail}, word prediction~\citep{Brown-2020-gpt3} or MRC problem~\cite{levy-2017-zero,Bragg-2021-flex} and use external resources from these tasks to pre-training the label semantic matching ability. However, the divergence between relation extraction and these tasks will inevitably undermine the performance. Besides, MRC and textual entailment architecture can only deal with one novel relation each time, which significantly increases the computational and storage cost of deploying such models in real-world scenarios. For few-shot RE, current methods mainly focus on summarizing better prototypes from a few illustrative instances~\cite{snell-17-proto}, or learning a model that can generalize to novel types within a few steps~\citep{finn-17-maml}. These approaches require few-shot examples to fine-tune or summarize prototypes, and therefore can not be directly applied to zero-shot RE. As a result, current relation extraction models can not be effectively and efficiently to apply to all low-shot RE settings.

\begin{figure*}[!ht]
    \setlength{\belowcaptionskip}{-0.4cm}
    \setlength{\belowcaptionskip}{-0.4cm}
    \centering
    \includegraphics[width=0.8\textwidth]{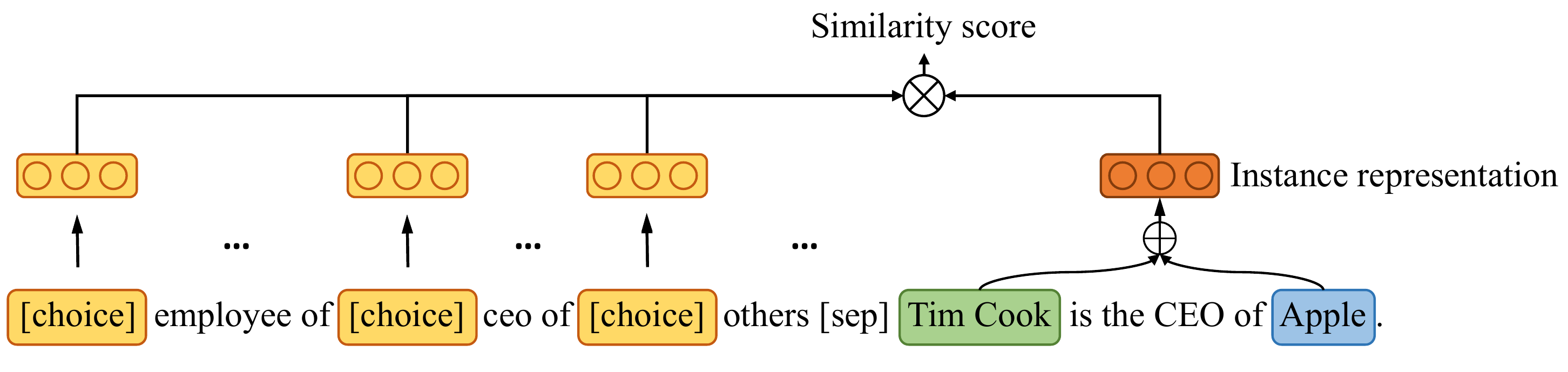}
    \caption{Illustration of our multi-choice matching networks~(MCMN).}
    \label{fig:mcmn}
\end{figure*}

In this paper, we propose to unify low-shot relation extraction by returning to the essence of relation extraction. Fundamentally, relation extraction can be viewed as a multiple choice task. Given two entities in context, a RE system needs to match the most appropriate relation -- or \emph{others} for none-of-the-above  -- from a set of pre-defined relation types. The information required to accomplish the multi-choice matching can be summarized from either the surface form of relation name or from few-shot instances. Motivated by this, we propose Multi-Choice Matching Network (MCMN) for unified low-shot RE, which is shown in Figure~\ref{fig:mcmn}. Specifically, MCMN converts all candidate relation descriptions into a multi-choice prompt. Then the input instance is concatenated with the multi-choice prompt and passes through a pre-trained encoder to obtain the semantic representations of the input instance and candidate relations.
Finally, MCMN conduct relation extraction by directly matching the relation representations and the instance representation.

To equip MCMN with both label semantic matching ability and instance semantic summarizing ability, we propose to pre-train MCMN via triplet-paraphrase meta pre-training, which contains the following two critical components: 1) a text-triple-text paraphrase module, which can generate large-scale pseudo relation extraction data to pre-train the label semantic matching ability of MCMN; 
2) a meta-learning style training algorithm, which enriches MCMN with instance semantic summarizing ability to quickly generalize across different relation extraction tasks. Specifically, given large-scale raw texts, triplet-paraphrase first extracts (subject, predicate, object) triplets via OpenIE~\citep{cui-2018-neural} toolkit. Then based on the extracted triplets, paraphrases of the original texts is generated 
using an RDF-to-Text generation model. In this way, we can obtain large-scale pseudo annotations by collecting the generated sentences and the predicate in the triples. Such corpus can be used to effectively pre-train the label semantic matching ability of MCMN by matching the paraphrases to the corresponding predicate. Furthermore, to enrich MCMN with the instance semantic summarizing ability, such pre-training is conducted in a meta-learning paradigm. That is, MCMN is asked to learn different relation extraction tasks at each iteration, so that the MCMN can not over-fit the pre-training corpus by directly memorizing specific target relations.

To evaluate our methods, we conduct experiments on three fairly different RE tasks: zero-shot RE, few-shot RE, and few-shot RE with none-of-the-above relation. Experiments show that the proposed method outperform previous methods on all these three tasks. Our source code is available at~\url{https://github.com/fc-liu/MCMN}.

The main contributions of this work are:
\begin{itemize}%[leftmargin=*]
    \item We propose MCMN, a unified architecture for low-shot relation extraction by fundamentally formulating relation extract using a multi-choice matching paradigm.
    \item We propose to pre-train MCMN with triplet-paraphrase meta training, which enriches MCMN with label semantic matching ability and instance semantic summarizing ability for both zero-shot RE and few-shot RE.
    \item We comprehensively study the performance of MCNN on three different relation extraction tasks, including zero-shot, few-shot, and few-shot with none-of-the-above relation extraction, where MCMN outperforms strong baseline models.
\end{itemize}

\section{Background}
In this section, we formulate relation extraction task and the low-shot RE settings including zero-shot RE and few-shot learning RE.

\paragraph{Relation Extraction.}
Suppose the input text $T=[t_1,t_2,...,tn]$ contains $n$ tokens, $e_1=[i,j]$ and $e_2=[k,l]$ indicate the entity pair spans, where $1\le i \le j, j< k\le l$ and $ l \le n$.
A relation instance is defined as $x=(T,e_1,e_2)$.
For example, the tuple (\emph{``Tim Cook is the CEO of Apple Inc.''},\emph{``Tim Cook''}, \emph{``Apple Inc.''}) is a relation instance.
The aim of relation extraction is to learn a mapping function: $f: x\boldsymbol{\to} y$, where $y$ is the relation class.
For example, we want mapping (\emph{``Tim Cook is the CEO of Apple Inc.''}, \ \emph{``Tim Cook''}, \ \emph{``Apple Inc.''}) to its relation class \emph{``CEO\_of''}. Traditional RE tasks typically pre-define the class space $Y$ and annotate a large set of instances to train the model. However, in real scenarios, the targeting relation types vary in different tasks, and most of the novel relations lack annotations, rendering the supervised paradigms inapplicable. In that regard, how to transfer models to novel tasks becomes critical.

\paragraph{Low-shot Relation Extraction.}
Low-shot relation extraction requires models to recognize novel relations with very few samples. There are two mainstream low-shot RE tasks, including:

\noindent \textbf{Zero-shot RE.} This task aims to conduct relation extraction without any annotated instance other than some external knowledge $z$~(or side information), such as relation descriptions. Models are supposed to transfer the knowledge and extract the targeting relation $y_t$ for input instance $x$ through only the external knowledge.

\noindent \textbf{Few-shot RE.} This task aims to conduct relation extraction with only a few annotated instances per novel relations. Each few-shot RE task contains a support set $S={S_1,...,S_N}$ for $N$ novel relations. And for relation $i$, $S_i={S_i^0,...,S_i^K}$ contains $K$ annotated instances. Models are supposed to learn to transfer the knowledge and extract the targeting relation $y_t$ for instance $x$ through the $N$-way $K$-shot support set.

\begin{figure*}[!ht]
    \centering
    \includegraphics[width=0.9\textwidth]{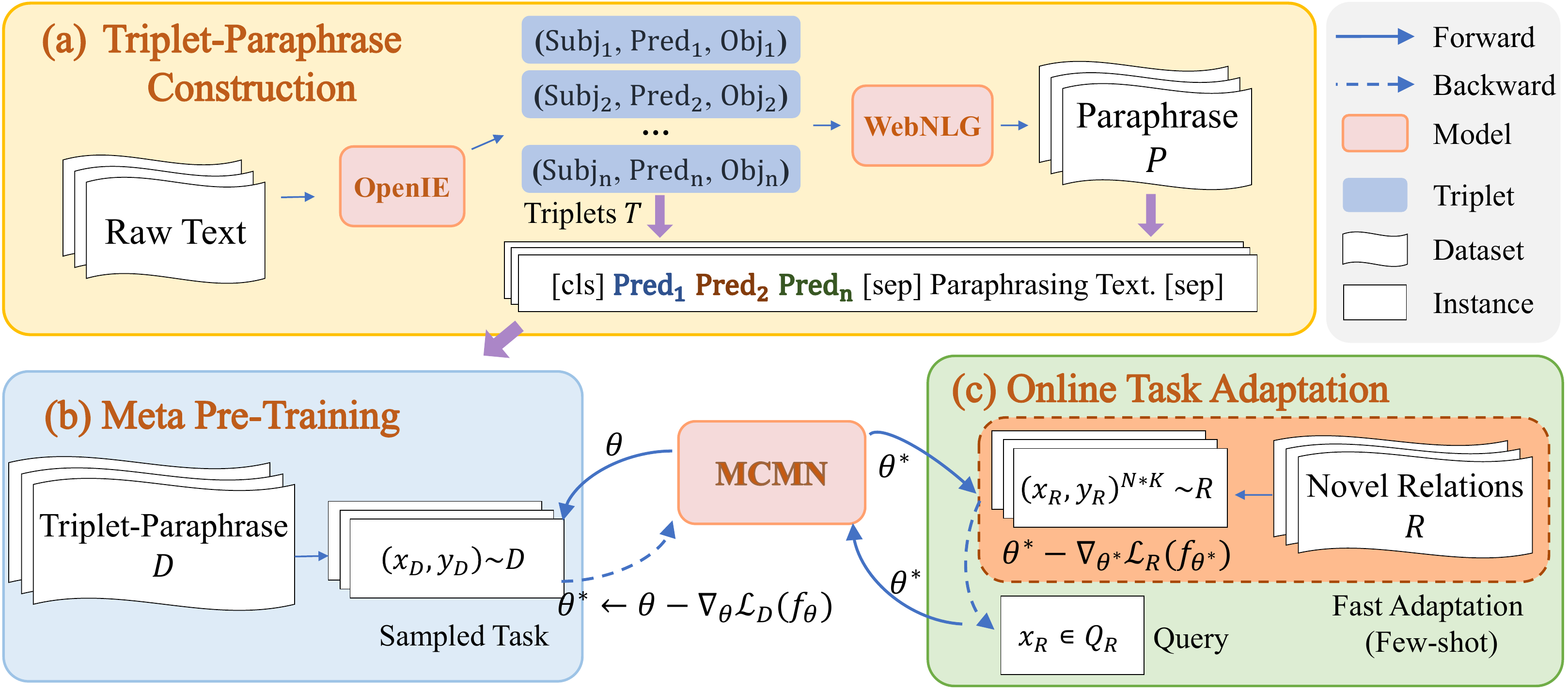}
    \caption{The framework of multi-choice matching training strategy. \textbf{(a) Triplet-Paraphrase Construction} conducts triplet-paraphrase pairs for meta training; \textbf{(b) Meta Training} on the triplet-paraphrase triplets; \textbf{(c) Online Task Meta-Training} performs an online meta-training for each test tasks.}
    \label{fig:pretrain}
\end{figure*}

\section{Multi-Choice Matching Networks}
In this section, we introduce our multi-choice matching networks~(MCMN). Different from previous unifying models, MCMN adopts a much more efficient and lightweight decoding module. Following are the detailed descriptions.

\subsection{Multi-choice Prompt}
Fundamentally, relation extraction can be viewed as a multiple choice task. Inspired by recent advances of prompt learning~\citep{Brown-2020-gpt3,schick-2021-prompt}, we construct a multi-choice prompt for each relation extraction task by directly concatenate all relation names or descriptions. Formally, the multi-choice prompts are in the following form:
\begin{center}
    [C] rel 1 [C] rel 2 ... [C] rel N
\end{center}
where [C] is the placeholder separator for the following relation. For example in Figure~\ref{fig:mcmn}, the target RE task contains three novel relations: \textit{employee\_of}, \textit{ceo\_of}, and \textit{others}, of which the relation descriptions are then concatenated altogether to form the multi-choice prompt ``[C] employee of [C] ceo of [C] others''. After obtaining the multi-choice prompt, we then feed it accompanied with the input sentence into the instance encoder, and the representations at separator [C] is regarded as the representation of its following relation.

\subsection{Instance Encoder}
Before instance encoding, we concatenate the multi-choice prompt with each input instance into a single sentence, and separate them with a [SEP] token. Besides, we follow \cite{soares-2019} and wrap the given entity pair with [e1], [/e1], [e2] and [/e2] respectively. For the example in Figure~\ref{fig:mcmn}, the entire input to encoder is: ``[CLS] [C] employee of [C] ceo of [C] others [SEP] [e1] Tim Cook [/e1] is the CEO of [e2] Apple [/e2] . [SEP]''. Then we encode the entire sentence $x$ through a Transformer~\citep{vaswani-2017-transformer} encoder:
\begin{equation}
    \textbf{h}_{[CLS]},\textbf{h}_{[C]},...,\textbf{h}_{[SEP]}=\mathcal{H}(x),
\end{equation}
where $\textbf{h}$ is the output embedding for each token in $x$, d is the dimension of hidden states. These token embeddings are then used for multi-choice matching and model prediction.

\subsection{Multi-choice Matching and Prediction}
The multi-choice matching module matches the input instance to the corresponding relation. For each relation type, we use hidden states of \textit{[C]} marker to represent each following relation:
\begin{equation}
    \textbf{h}_{rel_i}=\textbf{h}_{[C]_i},
\end{equation}
where $\textbf{h}_{rel_i}$ is the representation for relation $i$ and $\textbf{h}_{[C]_i}$ is the hidden state for the $i^{th}$ [C]  token.
For the input text, we simply average hidden states of \textit{[e1]} and \textit{[e2]} to obtain the instance representation $\textbf{X}$:
\begin{equation}
    \textbf{X} =avg(\textbf{h}_{[e1]}, \textbf{h}_{[e2]}).
\end{equation}
Then we perform matching operation between the instance and each relation:
\begin{equation}
    \mathcal{D}(x,y_i)=\|\textbf{X}-\textbf{h}_{rel_i} \|_2.
\end{equation}
In this equation, we adopt the Euclidean distance to measure the similarity, and the corresponding probability for each relation is:
\begin{equation}
    P(y_i\vert x;\theta)=\frac{\exp(-\mathcal{D}(x,y_i))}{\sum_{j=1}^{N}\exp(-\mathcal{D}(x,y_j))}.
\end{equation}
Finally, we choose the relation $\hat{y}$ with the maximal probability as the prediction:
\begin{equation}
    \hat{y} = \arg\max_{i}P(y_i|x;\theta).
\end{equation}

\subsection{Training Loss}
We adopt an end-to-end training manner by minimizing the following loss function:
\begin{equation}
    \setlength{\abovedisplayskip}{4pt}
    \setlength{\belowdisplayskip}{4pt}
    \mathcal{L}_{(x,y)}(\theta)=-\sum_{i = 1}^{N}\boldsymbol{I}(y_i)\log{P(y_i\vert x_i;\theta)},
    \label{eq:loss}
\end{equation}
where $\boldsymbol{I}(.)$ equals 1 if $y_i$ is the golden class, otherwise $\boldsymbol{I}(.)=0$. The three-period training process will be detailed described in the following section.

\section{Training Strategies for Multi-Choice Matching Networks}
As mentioned above, the required abilities for zero-shot and few-shot are different. In this paper, we propose triplet-paraphrase meta pre-training, which jointly learn the label semantic matching ability required by zero-shot RE and instance summarizing ability required by few-shot RE. Following is the detailed description of the pre-training framework.

\subsection{Triplet-Paraphrase Construction}
To endow the label semantic matching ability to MCMN, it is required to incorporate large-scale data of both relational sentences and relation types to pre-train the model. Unfortunately, the highly limited relation types in existing RE datasets may lead to overfitting on specific relations and impair the generalization of MCMN. In this paper, we propose the triplet-paraphrase to generate large-scale pre-training data for MCMN from raw texts. The overall procedure of triplet-paraphrase module is demonstrated in Figure~\ref{fig:pretrain}(a), which extracts predicates from large-scaled raw texts as the relation descriptions. Then we utilize the extracted relational triplets to generate paraphrase sentences for further multi-choice matching pre-training. The elaboration is presented below.

\paragraph{Relational Triplet Extraction.}
Most complete sentences contain at least one relational triplet, which includes the subject, predicate, and object. The predicate in a sentence corresponds to property or relation between the subject and object, which can be regarded as a concrete expression of one relationship. Therefore,  To extract large-scaled triplets from open domain texts, we use OpenIE model\footnote{\href{https://github.com/dair-iitd/OpenIE-standalone}{https://github.com/dair-iitd/OpenIE-standalone}} to extract on article collections of Wikipedia. Considering the example sentence: \textit{The service traces its history to an online service known as PlayNET.}
OpenIE model extracts all the possible triplets: (\textit{an online service, known as, PlayNET}) and (\textit{The service, traces, its history}).
We collect all extracted predicates from raw texts to represent the corresponding relations, preventing the models from overfitting specific relation types. These triplets are further used for paraphrase generation and pre-training.

\paragraph{Paraphrase Generation.}
One drawback of matching predicate as the relation is that the predicate extracted by OpenIE is commonly a span from current sentence, which may lead models to take the shortcut by directly matching through words co-occurrence. To eliminate this shortcut, we follow several recent works~\citep{agarwal-2021-tekgen,liu-2021-element} to generate paraphrase texts to match the predicate. Specifically, for extracted triplets, we first wrap them with special markers ``\textit{[H], [R], [T]}'' correspond to subject, predicate and object. Then we input the wrapped triplet texts to generate the paraphrase texts. In our implementation, we adopt T5\footnote{\href{https://github.com/UKPLab/plms-graph2text}{https://github.com/UKPLab/plms-graph2text}}~\citep{raffel-2020-t5} as the generator, and pre-train it on WebNLG dataset~\citep{gardent-2017-webnlg}. For example, we wrap (\textit{an online service, known as, PlayNET}) to ``[H] an online service [R] known as [T] PlayNET'' then generate the paraphrase text \textit{playnet is an online service.} After generating the paraphrase, we then match it to the corresponding predicate for pre-training.

\subsection{Triplet-Paraphrase Meta Pre-training}
Each instance in the pre-training batch contains the paraphrase text and the corresponding predicate span. In addition, as shown in Figure~\ref{fig:pretrain}(a), we concatenate all predicates in the current mini-batch as the multi-choice prompt and follow the training loss in Equation~\ref{eq:loss} to pre-train MCMN, where $\boldsymbol{I}(y_i)$ equals to 1 when $y_i$ is the corresponding predicate, otherwise, $\boldsymbol{I}(y_i)=0$. ·

\begin{algorithm}[!t]
    \caption{MCMN for Few-shot Prediction}
    \label{alg1}
    \begin{algorithmic}[1]
        \REQUIRE $n$: fine-tuning epochs in online period
        \REQUIRE $\theta^*$: meta-learned model parameters
        \REQUIRE $\mathcal{S}$: support set, $x_q$: query instance
        \REQUIRE $\alpha$: learning rate
        \STATE $\theta'=\theta^*$ \# save original model
        \FOR{epoch\ in\ \textbf{range}(n)}
        \STATE \# compute loss of the support set:
        \STATE $\mathcal{L_{\mathcal{S}}}=\mathbb{E}_{(x,y)\in \mathcal{S}}\mathcal{L}_{(x,y)}(\theta^*)$
        \STATE \# update model parameters:
        \STATE $\theta^*\leftarrow \theta^*-\alpha\nabla_{\theta^*}\mathcal{L_{\mathcal{S}}}$
        \ENDFOR
        \STATE $y=f_{\theta^*}(x_q)$ \# predict the query instance
        \STATE $\theta^*=\theta'$ \# restore the original model
        \RETURN $y$
    \end{algorithmic}
\end{algorithm}

\begin{table*}[htbp]
    \setlength{\belowcaptionskip}{-0.4cm}
    \centering
    \resizebox{0.95\textwidth}{!}{
        \begin{tabular}{l|cccc|c}
            \toprule
            \cmidrule{1-5}    \multirow{2}[4]{*}{\textbf{Model}} & \multicolumn{2}{c|}{\textbf{Zero-shot}}          & \multicolumn{2}{c|}{\textbf{Few-shot}} & \multirow{2}[4]{*}{\textbf{Avg.}}                                              \\
            \cmidrule{2-5}                                       & \textbf{Acc.$\pm$ci.}                            & \multicolumn{1}{c|}{\textbf{Std.}}     & \textbf{Acc.$\pm$ci.}             & \textbf{Std.}                              \\
            \midrule
            UniFew~\citep{Bragg-2021-flex}                       & 52.5$\pm$ 2.0                                    & \multicolumn{1}{c|}{9.7}               & 79.2$\pm$ 1.5                     & 7.5                       & 65.9           \\
            UniFew-meta~\citep{Bragg-2021-flex}                  & 79.4 $\pm$ 1.9                                   & \multicolumn{1}{c|}{9.2}               & 87.2$\pm$ 1.2                     & 5.7                       & 83.3           \\
            \midrule
            MCMN w. Pre-train Only & 66.6$\pm$ 1.7 & \multicolumn{1}{c|}{8.7} & 74.4$\pm$ 1.5 & 7.6 & 70.5 \\
            MCMN                                                & \textbf{82.9}$\pm$1.3                            & \multicolumn{1}{c|}{\textbf{6.6}}      & \textbf{87.4}$\pm$1.2             & \textbf{5.6}              & \textbf{85.1}  \\
            \bottomrule
            \multirow{2}[4]{*}{\textbf{Model}}                   & \multicolumn{4}{c|}{\textbf{Few-shot with NOTA}} & \multirow{2}[4]{*}{\textbf{Avg.}}                                                                                       \\
            \cmidrule{2-5}                                       & \textbf{5-way 1-shot 0.15}                       & \textbf{5-way 5-shot 0.15}             & \textbf{5-way 1-shot 0.5}         & \textbf{5-way 5-shot 0.5}                  \\
            \midrule
            Proto~(CNN)~\citep{gao-2019-fewrel2} & 60.59                                            & 77.79                                  & 40.00                             & 61.66            & 60.01          \\
            Proto~(BERT)~\citep{gao-2019-fewrel2} & 70.02                                            & 83.79                                  & 45.94                             & 75.21                     & 68.74          \\
            Bert-Pair~\citep{gao-2019-fewrel2}                  & 77.67                                            & 84.19                                  & 80.31                             & \textbf{86.06}                     & 82.06          \\
            $2^{nd}$ on Leaderboard~(anonymous)                  & 79.53                                            & 86.31                                  & 79.99                             &81.92            & 81.94          \\
            $3^{rd}$ on Leaderboard~(anonymous)                  & 67.97                                            & 81.94                                  & 74.85                             & 78.12                     & 75.72          \\
            \midrule
            MCMN~($1^{st}$ on Leaderboard)                        & \textbf{88.40}                                   & \textbf{89.91}                         & \textbf{84.56}                   & 85.32                     & \textbf{87.05} \\
            % Our~(0-shot)                                         & 69.31                                            & 69.25                                  & 77.15                             & 76.45                     & 73.04          \\
            \bottomrule
        \end{tabular}%
    }
    \caption{Results~(\%) on zero-shot, few-shot, and few-shot with NOTA RE tasks. We report accuracy~(Acc.), confidence interval~(ci.), and standard deviation~(Std., lower is better) for zero- and few-shot RE, and only accuracy for few-shot with NOTA task.}
    \label{exp:main}
\end{table*}%

\section{Experiments}
\subsection{Datasets and Task Settings}
We conduct experiments on three low-shot relation extraction tasks: zero-shot RE~\citep{Bragg-2021-flex}, few-shot RE~\cite{Bragg-2021-flex} and the more challenging few-shot RE with none-of-the-above~(NOTA)~\citep{gao-2019-fewrel2}. These tasks are all conducted based on FewRel dataset~\citep{han-2018-fewrel}, which is constructed through distantly aligning WikiData triplets to Wikipedia articles. In total, FewRel dataset consists of 100 relation types and 700 instances per type. Standard FewRel settings adopt a split of 64/16/20 fraction corresponding to train/validation/test set, where the train and validation sets are publicly accessible while the test set is not. Following are the detailed settings for each evaluation task.
\paragraph{Zero- and Few-shot Relation Extraction Settings.}
We follow the standard Flex benchmark settings, which separate the train and validation sets from FewRel into a train set of 65 relations, a validation set of 5 relations and a test set of 10 relations. The test tasks are sampled and processed through the FLEX official toolkit~\footnote{\href{https://github.com/allenai/flex}{https://github.com/allenai/flex}}.

\paragraph{Few-shot RE with NOTA Relation Settings.}
A drawback of conventional few-shot RE tasks is that they neglect the existence of other relations, that is all query instances are assumed to express one of the given relations in the support set. \citet{gao-2019-fewrel2} point out this problem and add the \textit{``none-of-the-above''}~(NOTA) relation to consider the situation where query instance does not express any of the given relations. In our experiment, we follow the default settings of FewRel benchmark and evaluate our methods on 5-way 1/5-shot tasks with a 15\% or 50\% NOTA rate.

\subsection{Baseline and Evaluation Metrics}
\paragraph{Baseline Methods.} For zero-shot and few-shot RE tasks, we compare our model with \textbf{UniFew}~\citep{Bragg-2021-flex}, a unified few-shot learning model based on T5~\citep{raffel-2020-t5}. This model converts each few-shot classification task into a machine reading comprehension format and predicts the results through generation. With a pre-training period on large-scaled MRC data, this model reaches strong performance on both zero- and few-shot tasks. For few-shot RE with NOTA relation task, we compare our model with \textbf{Bert-Pair}~\citep{gao-2019-fewrel2}, an instance-pair matching framework for few-shot RE tasks. This model computes a similarity and a dissimilarity score simultaneously between query instance and each support instance, then aggregates the similarity score for each relation and dissimilarity score for NOTA relation. And the results of \textbf{CNN} and \textbf{BERT} based prototypical networks from \citet{gao-2019-fewrel2} are also reported.
\paragraph{Evaluation Metrics.} For zero-shot and few-shot RE tasks, we follow FLEX benchmark and report the accuracy, confidence interval and standard deviation correspondingly. All these results reported are from the official Flex toolkits. For few-shot RE with NOTA relation task, we follow FewRel 2.0 benchmark and report the corresponding accuracy for four different settings.

\subsection{Hyperparameters and Implementation Details}
In the triplet-paraphrase construction period, we extract relation triplets from articles in Wikipedia and generate the counterpart paraphrase texts. Overall, we generate about one million triplet and paraphrase text pairs. In triplet-paraphrase meta-training periods, we use a learning rate of 5e-6, weight decay of 1e-6, dropout rate of 0.5, and a linear learning schedule with a 0.95 weight decay. In the online task meta-training period, we use learning rate of 5e-6, and the adaptation epoch $n$ of 1 or 2 for FewRel NOTA tasks, epochs of 45 for FLEX tasks, while keep other hyperparameters the same. We use RoBERTa-large~\citep{liu-2019-roberta} to initialize our model. Furthermore, to better endow the low-shot capability to our model, we adopt annotated FewRel data~\citep{han-2018-fewrel} as an additionally supervised meta-training procedure.

\subsection{Overall Results}
Table~\ref{exp:main} shows the overall results on three different RE tasks. From this table, we can see that:
\begin{itemize}[leftmargin=*]
    \item \textbf{MCMN with triplet-paraphrase pre-training outperforms previous methods in all three RE tasks and achieve state-of-the-art performance.} Compared with the strong baseline methods, MCMN achieves remarkable performance improvements.  In zero-shot and few-shot RE tasks, MCNN with triplet paraphrase pre-training outperforms the baseline methods by at least 1.8\% in average. In few-shot RE with NOTA task, our method outperforms previous best method by at least 4.99\% in average and achieve the best performance in the leaderboard.
    \item \textbf{Our triplet-paraphrase pre-training achieves promising results on low-shot RE tasks.} Comparing with other pre-training strategies such as UniFew model pre-trained with large annotated MRC datasets, triplet-paraphrase pre-training achieves much better performance on zero-shot RE tasks. Besides, triplet-paraphrase can further enhance  MCMN to achieves the new state-of-the-art results on all three low-shot RE tasks with supervised meta-training procedure, which are detailed analyzed in the next section.
    \item \textbf{MCMN performs more robust than previous methods.} In zero-shot and few-shot tasks, our methods perform a lower standard deviation and more shallow confidence interval than baseline methods, which means the prediction of our methods is more stable across different tasks. 
\end{itemize}

\subsection{Detailed Analysis}
In this section, we conduct several experiments for in-depth analysis of our methods.

\begin{table}[t]
    \setlength{\belowcaptionskip}{-0.4cm}
    \centering
    \resizebox*{0.45\textwidth}{!}{
    \begin{tabular}{l|cc|cc|c}
        \toprule
        \multirow{2}[4]{*}{Model} & \multicolumn{2}{c|}{Zero-shot} & \multicolumn{2}{c|}{Few-shot}               &
        \multirow{2}[4]{*}{\textbf{Avg.}}  \\
        \cmidrule{2-5}            & acc                            & std.                         & acc. & std. \\
        \midrule
        RoBERTa             & 15.6                           & 5.1                          & 21.4 & 7.3 & 18.5  \\
        Triplet-Para Pre-train                & 66.6                           & 8.7                          & 74.4 & 7.6 & 70.5 \\
        MCMN w/o Pre-train                    & 81.0                           & 6.7                          & 85.3 & 5.7 & 83.2 \\
        \midrule
        MCMN                  & 82.9                           & 6.6                          & 87.4 & 5.6 & 85.1 \\
        \bottomrule
    \end{tabular}%
    }
    \caption{Ablation study results~(\%) of our methods on FLEX RE benchmark, Sup. Meta stands for supervised meta training.}
    \label{exp:pretrain}%
\end{table}%

\paragraph{Ablation Studies on Zero- and Few-shot RE Tasks.}
To evaluate the effect of each part of our methods on zero- and few-shot RE tasks, we conduct separate experiments on triplet-paraphrase pre-training, MCMN and MCMN without triplet-paraphrase pre-training on Flex test set. As shown in Table~\ref{exp:pretrain}, we can see that the pure triplet-paraphrase pre-training model outperforms RoBERTa-large model with a remarkable margin as well as leverages the MCMN model with an improvement of at least 1.9\% compared with MCMN without triplet-paraphrase pre-training on both zero-shot and few-shot settings. These results demonstrate that triplet-paraphrase pre-training method can significantly improve the generalization and performance of our model, and the framework of multi-choice matching network is quite applicable in low-shot RE tasks. 
Besides, we notice the performance of pure triplet-paraphrase pre-training model is lower than MCMN without triplet-paraphrase pre-training. To study this issue, we analyze the triplet-paraphrase data, and find that many of the generated texts still consist of words in predicates, though the expression is quite different from the original sentences. This may still lead to the shortcut learning problem. On top of that, the expression of predicates is much different from the relation name, and the negative predicates are much easier to distinguish than the real test cases. These issues altogether result in poor performance. Fortunately, the triplet-paraphrase pre-training period can properly initialize MCMN and leverage the final performance.

\begin{table*}[htbp]
    \setlength{\belowcaptionskip}{-0.4cm}
    \centering
    \resizebox{0.95\textwidth}{!}{
        \begin{tabular}{l|cccc|c}
            \toprule
            \multirow{2}[4]{*}{\textbf{Model}}                   & \multicolumn{4}{c|}{\textbf{Few-shot with NOTA}} & \multirow{2}[4]{*}{\textbf{Avg.}}                                                                                       \\
            \cmidrule{2-5}                                       & \textbf{5-way 1-shot 0.15}                       & \textbf{5-way 5-shot 0.15}             & \textbf{5-way 1-shot 0.5}         & \textbf{5-way 5-shot 0.5}                  \\
            \midrule
           RoBERTa & 27.37 & 27.88 & 16.38 & 16.50 & 22.03 \\
           Triplet-Para Pre-train & 69.00                                            & 70.59                                  & 43.99                             & 43.66            & 56.81          \\
           MCMN w/o. Pre-train & 87.89                                            & 90.36                                  & 83.22                             & 83.10                     & 86.14          \\
            MCMN                        & 88.40                                   & 89.91                         & 84.56                   & 85.32                     & 87.05 \\

            \midrule
            MCMN w/o. Pre-train~(\textbf{0}-shot)                                         & 83.08                                            & 84.10                                  & 83.61                             & 83.45                     & 83.56          \\
            MCMN~(\textbf{0}-shot) & 85.11 & 85.45 & 82.72 & 82.16 & 83.86 \\
            \bottomrule
        \end{tabular}%
    }
    \caption{Ablation study results~(\%) of MCMN on FewRel NOTA benchmark. Triplet-Para corresponds to triplet-paraphrase.}
    \label{exp:nota}
\end{table*}%

\paragraph{Ablation Studies on Few-shot NOTA RE tasks.}
We also conduct detailed analyses of our methods in few-shot NOTA RE tasks. As shown in Table~\ref{exp:nota}, the pure triplet-paraphrase pre-trained model can also boost the performance of roberta-large initialized model and leverage the supervised meta-trained MCMN by at least 0.9\% in average. Although we do not consider the NOTA relation in the triplet-paraphrase pre-training period, this period can also contribute to the further supervised meta-training period, which indicates that the matching-pattern learned in triplet-paraphrase pre-training period is generalized and robust to down-stream tasks.
Besides, we notice that in NA rate of 0.5 tasks, the pure triplet-paraphrase pre-trained model suffers from serious performance drops. This may be caused by the large proportion of negative instances in test tasks. Fortunately, this issue can be alleviated by the online adaptation period.

\paragraph{Zero-shot NOTA RE tasks.}
This experiment studies the zero-shot performance of our methods on FewRel NOTA tasks. From Table~\ref{exp:nota}, we surprisingly found that our methods also outperform previous state-of-the-art few-shot NOTA models even in zero-shot conditions. This also indicates that our methods are effective in low-shot RE tasks and are robust enough across different settings.

\paragraph{Computing Efficiency of Multi-Choice Matching Networks.}
This experiment compares the computing efficiency of our method with MRC-based method. Each model is tested on the Flex test set, including both zero-shot and few-shot RE tasks. 
Models in zero-shot setting only need inference while both models in few-shot setting require fine-tuning on the support set which involves time-consuming back-propagation operations. For fair comparison, we use a single TITAN RTX GPU for each model and keep other computing environments the same. As a result, UniFew takes \textbf{647} minutes~(more than 10 hours) to finish the test prediction, while our method takes about \textbf{80} minutes to obtain the results in Table~\ref{exp:main}, which improves the speed of roughly an order of magnitude.   The main reason for such an efficiency discrepancy is that UniFew, as a generative model, involves an auto-regressive decoder to generate the results, whereas our method directly matches the relation and instance representations to give the results.

\section{Related Works}
\paragraph{Relation Extraction.} Recent success of supervised relation extraction methods~\citep{zeng-2014,zhou-2016-lstm-rc} heavily depends on large amount of annotated data. However, the bottleneck on data annotation severely limits the adaptation of these supervised methods to real scenarios. Recent works reply to this dilemma from the perspective of low-shot learning, which mainly focuses on zero- and few-shot RE tasks. In this work, we shed light on three representative sub-fields tasks, including zero-shot RE, few-shot RE and few-shot RE with NOTA relation to evaluate our methods

\paragraph{Zero-shot Relation Extraction.}\citet{levy-2017-zero} firstly introduce the zero-shot relation extraction task and adjust the machine reading comprehension~(MRC)-based paradigm for it. Following this line, other MRC-based methods have been proposed~\citep{cetoli-2020-exploring,Bragg-2021-flex}. Another paradigm for zero-shot RE is matching-based~\citep{zero-shot-match}, which falls into the text-entailment-based methods~\citep{obamuyide-2018-zero,sainz-2021-entail}, and the representation matching-based methods~\citep{chen-zs-bert,dong-2021-mapre}. Text-entailment-based methods concatenate the relation description with the input sentence to assess whether they entail the same semantic relationship; Representation matching-based methods separately encode the relation and instance into the same semantic space but are not capable of handling the NOTA relation.
\paragraph{Few-shot Relation Extraction.} \citet{han-2018-fewrel} firstly propose the few-shot relation extraction task and adopt several meta-learning methods~\citep{munkhdalai-17-metanet,snell-17-proto,garcia-2018-fsgnn,mishra-2018-snail} for it. Recent works on few-shot RE mostly centers around the metric-based methods~\citep{vinyal-16}, such as prototype-based methods~\citep{soares-2019,ye-2019-MLMAN,gao-2019-hybrid} and meta learning-based methods~\citep{finn-17-maml}. Besides, \citet{gao-2019-fewrel2} extend the FewRel challenge with few-shot domain adaptation~(DA) and none-of-the-above~(NOTA) tasks, which are more challenging and closer to real-world application.
\paragraph{Few-shot RE with NOTA.} Although NOTA relation is common in conventional supervised RE tasks~\citep{zhang-2017-tacred}, it is quite different in few-shot scenarios due to the label inconsistency problem. As an example, consider an instance that expresses relation r. In task A, relation r is not included in the support set, and thus model learns the semantic mapping between this instance and the NOTA relation. But in another task B where relation r is included in the support set, the model learned from task A may continue to match this instance to NOTA relation. Because of the difficulty, attempts to resolve this problem are scarce. To the best of our knowledge, Bert-Pair \citep{gao-2019-fewrel2} is the only public method for this task, and our work is the first method to unify the zero-shot, few-shot and few-shot with NOTA tasks.

\section{Conclusions}
In this paper, we propose Multi-Choice Matching Networks to unify low-shot relation extraction. MCMN introduces a multi-choice prompt to formulate relation extraction as in a multi-choice paradigm. To equip MCMN with different zero-shot and few-shot abilities, we propose the triplet-paraphrase meta pre-training, which leverages triplet paraphrase to pre-train zero-shot label matching ability and uses meta-learning paradigm to learn few-shot instance summarizing ability. Experimental results on three different RE tasks show MCMN outperforms strong baseline models by large margins.

\section*{Acknowledgements}
We thank all reviewers for their insightful suggestions. Moreover, this research work is supported by the Strategic Priority Research Program of Chinese Academy of Sciences under Grant No. XDA27020200, and the National Natural Science Foundation of China under Grants no. 62106251 and 62076233.

\section*{Ethics Consideration}
This paper has no particular ethic consideration.

%\newpage
\bibliography{custom}
\bibliographystyle{acl_natbib}

\end{document}